\documentclass[journal]{IEEEtran}
\usepackage{amsmath,amssymb}             
\usepackage{amsfonts}
\def\argmax{\operatornamewithlimits{arg\,max}}
\def\argmin{\operatornamewithlimits{arg\,min}}
\usepackage{graphicx}
\usepackage{algorithm}
\usepackage{algpseudocode}
\usepackage{cite}
\usepackage{comment}
\usepackage{multirow}
\newcommand{\REV}[1]{{\textcolor{black}{#1}}}

\usepackage[table,xcdraw]{xcolor}

\usepackage{ragged2e}
\begin{document}
\title{Weakly-Supervised Learning of Metric Aggregations for Deformable Image Registration}
%
%
%

\author{Enzo Ferrante\thanks{E. Ferrante is affiliated with the Research institute for signals, systems and computational intelligence, sinc(i), FICH-UNL/CONICET. (Santa Fe, Argentina). Mail: eferrante@sinc.unl.edu.ar}, Puneet K. Dokania\thanks{P. K. Dokania is affiliated with the University of Oxford (United Kingdom).}, Rafael Marini Silva, Nikos Paragios~\IEEEmembership{Fellow, IEEE}\thanks{R. Marini Silva and N. Paragios are affiliated with TheraPanacea and with the Center for Visual Computing, CentraleSupelec, Universite Paris-Saclay (Paris, France}} 

%
%

\markboth{IEEE Journal of Biomedical and Health Informatics}%
{Shell \MakeLowercase{\textit{et al.}}: Bare Demo of IEEEtran.cls for IEEE Journals}
%

\maketitle

\begin{abstract}
Deformable registration has been one of the pillars of biomedical image computing. Conventional approaches refer to the definition of a similarity criterion that, once endowed with a deformation model and a smoothness constraint, determines the optimal transformation to align two given images. The definition of this metric function is among the most critical aspects of the registration process. We argue that incorporating semantic information (in the form of anatomical segmentation maps) into the registration process will further improve the accuracy of the results. In this paper, we propose a novel weakly supervised approach to learn domain specific aggregations of conventional metrics using anatomical segmentations. This combination is learned using latent structured support vector machines (LSSVM). The learned matching criterion is integrated within a metric free optimization framework based on graphical models, resulting in a multi-metric algorithm endowed with a spatially varying similarity metric function conditioned on the anatomical structures. We provide extensive evaluation on three different datasets of CT and MRI images, showing that learned multi-metric registration outperforms single-metric approaches based on conventional similarity measures.
\end{abstract}

%
\IEEEpeerreviewmaketitle

\section{Introduction}
%
%
%
%

Image registration deals with the alignment of images of a same object coming from different devices, moments or viewpoints. When the observed images are mapped through a nonlinear dense transformation or a spatially varying deformation model, we refer to this problem as deformable image registration  (DIR) \cite{Sotiras2013}. Inspired by Horn and Shunk \cite{Horn1980} and the seminal work by Lucas and Kanade on vector flow estimation \cite{Lucas1981}, the research communities of computer vision and medical image analysis have made major efforts towards developing more accurate and efficient DIR methods. Since then, image registration has been modeled through different approaches, ranging from diffusion equations to probabilistic graphical models \cite{Paragios2016}. The majority of these approaches posed image registration as an optimization problem, following:

\begin{equation}
 \hat{T} = \argmin_{T} \mathcal{M}(I \circ T, J) + \mathcal{R}(T),
 \label{eq:genericFormulationRegistration}
\end{equation}

\noindent where $\mathcal{M}$ represents the data term or (dis)similarity criterion (measuring the (dis)similarity between the warped source image $I \circ T$ and the target image $J$) and $\mathcal{R}$ the regularization term (used to impose geometric consistency and ensure realistic deformations). Here $\hat{T}$ indicates the optimal transformation. 

The definition of the similarity criterion $\mathcal{M}$ is among the most critical components of the registration process, since it specifies whether the source and target images are correctly aligned or not. It has been empirically shown that the data terms in the energy function have great influence on the accuracy of the solution to the registration problem. Normally, a data term is the function of a metric such as the sum of absolute differences (SAD) of the intensities, the mutual information (MI) between the images or their normalized cross correlation, just to name a few (we refer the reader to \cite{Sotiras2013} for a complete list of standard similarity measures). A particular data term is thus chosen based on the application. We argue that instead of using one metric, the locally adaptive and content-specific combination of different metrics can further improve the accuracy of the registration task, in particular, in the presence of semantic labels which then make the registration a domain specific problem. In this paper, we present a method that aims at learning context-specific matching criteria as a weighted aggregation of standard similarity measures. We adopt a weakly-supervised approach, in the sense that we do not use ground-truth annotations in the form of deformation fields. Instead, we employ segmentation maps of anatomical structures as a proxy to impute such latent deformation fields. Specifically, we use the latent structured support vector machine (lssvm)~\cite{Yu2009} so that the optimal combinations of standard similarity measures can be learned from a few examples.

\subsection{State-of-the-art on Learning Similarity Measures for Medical Image Registration}
During the last years, efforts have been made towards learning similarity measures from data, most of them by posing this problem as a metric learning problem \cite{Yang2006}. Metric learning aims at determining a mapping that will bring all data points within the same classes close, while separating all the data points from different classes. From this general definition, the meaning of metric learning in the context of image registration can be simply interpreted as learning a domain specific matching criterion that allows the comparison of any two given image modalities. In \cite{Yang2006}, metric learning methods were classified as supervised or unsupervised.

Supervised methods require a set of annotated data (i.e. some sort of similarity information, like a full scalar distance between samples, a ranking or some proximity of some samples to others or just a separation between samples that are deemed similar and dissimilar \cite{Michel2014}). Several image registration methods using supervised metric learning can be found in the recent literature. \cite{Lee2009} derived a rigid multi-modality registration algorithm where the similarity measure is learned in a discriminative manner, such that the target and correctly deformed source image receive high similarity scores using a max-margin approach known as structured prediction. In a different but still supervised setting, \cite{Bronstein2010} proposed the use of sensitive hashing to learn a multi-modality distance metric. The training set involved pairs of perfectly aligned images and consisted of a collection of positive and negative patch pairs. Such a metric learning approach was plugged into the standard graph-based deformable registration framework in \cite{Michel2011}. In \cite{Toga2008}, instead of requiring pre-aligned images, the training set consisted of non-aligned images with manually annotated patch pairs (landmarks). All these methods require ground truth data in the form of strong correspondences (patches, landmarks or dense deformation fields), which is extremely difficult to obtain in real clinical data. Instead, the method we propose in this work requires segmentation masks which are easier to generate \REV{(e.g using deep learning based segmentation strategies like \cite{Shakeri2016})}. More recently, approaches based on deep learning started to gain popularity. The work by \cite{Zagoruyko2015} proved that similarity measures can be learned from annotated image patches using convolutional neural networks (CNN), and discussed different models that could be used to that end. In \cite{Simonovsky2016}, the best model presented by Zagoruyko was used as a similarity measure in a standard continuous framework for image registration. The main limitation of CNN-based approaches to metric learning for image registration is that they require huge amounts of data to learn meaningful correspondences. 

Unsupervised learning methods for image registration have also been studied. \cite{Wachinger2010} proposed to apply manifold learning (through Laplacian eigenmaps) to learn structural representations of multimodal images. In \cite{Ou2011} the notion of mutual saliency was considered to locally adapt the metric function involving a weighted sum over a large space of features. A different approach based on unsupervised deep learning was proposed by \cite{Wu2013}. The composition of a basis of filters was learned to effectively represent the observed image patches. Another unsupervised deep learning approach was proposed in \cite{yoo2017}. The authors combined a spatial transformer \cite{jaderberg2015spatial} network with a convolutional autoencoder to perform unsupervised feature learning for robust serial section electron microscopy (ssEM) image alignment. Different from our weakly-supervised approach, these unsupervised models do not incorporate any type of semantic information about anatomy into the registration process.

Most similar to our work is the recent method by \cite{hu2017label}, where a label-driven weakly supervised approach based on CNNs is proposed. Like in \cite{jaderberg2015spatial}, they learn the complete registration process by means of a spatial transformer layer, but additionally, they incorporate anatomical segmentations into the loss function. The main difference with our work is that they learn the complete registration process from scratch (since the learned model directly predicts the optimal deformation field), and therefore it is not possible to plug the trained model as a similarity measure in a standard registration framework.

\subsection{Our Contribution: Label-driven Weakly-supervised Learning of Standard Metric Aggregations}
In this work, we focus on cases where the ground-truth is given in the form of segmentation masks, which is more common in real scenarios than availability of dense deformation fields or pre-aligned images. In a number of applications, segmentation maps might be available in one of the volumes to be registered (e.g. adaptive radiotherapy, patient follow up exams or multi-atlas segmentation). Human anatomy is domain specific (deformations as well) and depending on the organ to be imaged one can expect different adequacy between the local structure and different mono or multimodal metrics (which somehow reflect the observed tissue properties). Hence, the available segmentation maps could be used to locally suggest efficient similarity measures depending on the organ. We propose a novel theoretical framework where domain specific optimal combination of standard metrics is derived according to the clinical task / observed images. The degrees of freedom of the learning procedure are the class  weight combination of the different metrics. 


The idea of combining different similarity measures (not necessarily considering them as domain or content-specific) has already been explored. In \cite{Cifor2012} it was shown that multichannel registration produces more robust registration results when compared to using the features individually. In this case, they used gray intensity value, phase congruency and local phase as features. 
In a posterior work, \cite{Cifor2013a} proposed a methodology that does not require to explicitly weight the features, by estimating different deformation fields from each feature independently, and then composing them into a final diffeomorphic transformation. Such a strategy produces multiple deformation models (as many as the number of metrics) that in general are locally inconsistent. Therefore, their combination will not be trivial and, in the general case, not anatomical meaningful.

In deformable registration, it is crucial to choose the right relative weighting between the different metrics and the pairwise smoothness term. One naive way would be to choose these relative weights by cross validation (or hand tuning) over the parameters. Clearly, this approach quickly becomes infeasible as the number of metrics increases while it can be applied only at global scale and cannot accommodate local anatomical differences. In order to circumvent the above mentioned problem, we would like to learn the relative weights from a given training dataset using a learning framework. Our method is similar to that of Tang et al. \cite{tang2012locally}, which generates a vector weight map that determines at each spatial location the relative importance of each constituent of the overall metric. However, the proposed learning strategy still requires ground truth data in the form of correspondences (pre-registered images) which is not necessary in our case. We propose a novel discriminative learning framework, based on the well known structured support vector machines ({\sc ssvm})~\cite{Taskar2003, Tsochantaridis2004}, to learn the relative weights (or the parameters) from weak annotations. The {\sc ssvm} and its extension to latent models {\sc lssvm}~\cite{Yu2009} have received considerable attention in the recent years for parameter learning.

Our focus is mainly on the 3D to 3D deformable registration problem where the input and the output images are 3D volumes. However, the same framework can be trivially used for other registration problems as well, as 2D to 2D, or even for slice-to-volume registration \cite{ferrante2017slice}. One of the key issues we face in all these problems is that the ground truth deformations are not known. This leads us to adopt a weakly supervised learning framework where we treat the ground truth deformations as the latent variables. We model the latent variable imputation problem as the deformable registration problem with additional constraints. These constraints ensure that when the latent deformations are applied to the source image, the deformed source image is maximally aligned with the target image. The alignment accuracy is measured based on a loss function. 

Our learning framework, similar to the {\sc lssvm}, is a special case of non-convex optimization problems, known as the difference of convex functions. The local optimum or the saddle point of such non-convex function can be obtained efficiently using the well known {\sc cccp} algorithm~\cite{Yuille2003}. We demonstrate the efficacy of our framework using three challenging medical imaging datasets. A preliminary version of this work has been reported in the International Workshop on Machine Learning in Medical Imaging (MLMI 2017) \cite{ferrante2017deformable}. In this special issue of JBHI, we present an invited extended version with a more complete and updated state-of-the-art section, detailed method explanation and additional figures and discussions.

\section{Multi-Metric Deformable Registration}
\label{sec:regProblem}
\begin{figure}[t!]
  \centering
  \includegraphics[width=0.45\textwidth]{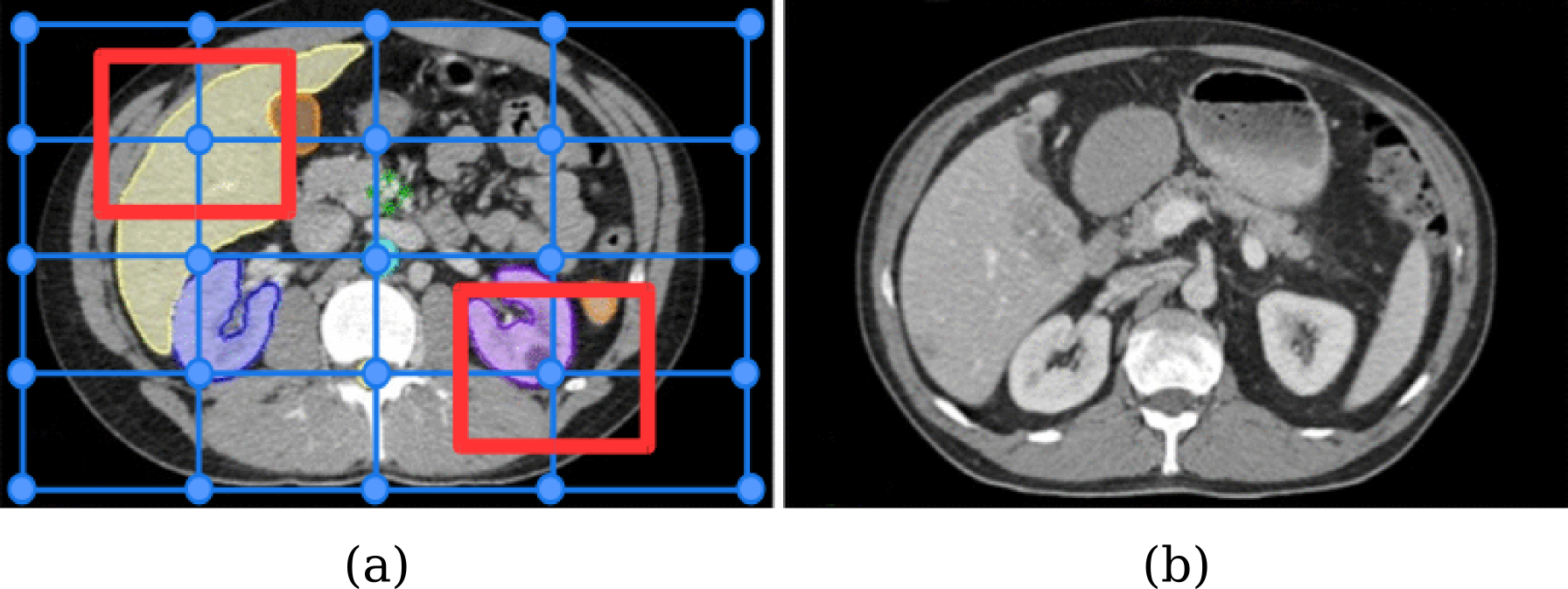}
  \caption{The multi-metric deformable registration algorithm uses a different aggregation of metrics depending on the context, which is determined by the dominant class in the corresponding source image support area. In the example, we can observe that the liver (in yellow) and the kidney (in violet) are the dominant classes for the two highlighted control points. Note that, during training, both source (a) and target (b) semantic labels are required to compute the loss function $\Delta$. However, at test time, we only require semantic labels for the source image (a) to choose the dominant class.} 
  \label{fig:dominantClass}
\end{figure}

Let us assume that we are given a source $3D$ volume (or image) $I$, source $3D$ segmentation mask $S^I$, and the target $3D$ volume (or image) $J$. The size of the segmentation mask is the same as that of the corresponding image. The segmentation mask is formed by the elements (or voxels) $s_k \in \mathcal{C}$, where $\mathcal{C}$ is the set of classes. \REV{Every class $c \in \mathcal{C}$ is associated to a different anatomical structure (organs, tissues, etc) as depicted in Figure \ref{fig:dominantClass}.} Without loss of generality, we assume that the elements in the class set $\mathcal{C}$ are the discrete variables starting from one. 

The registration problem is formulated following \cite{Glocker2008} using a \REV{Markov Random Field (MRF)} that consists of a regular grid graph $\mathcal{G}=\langle \REV{V},\REV{E} \rangle$, where \REV{$V$} is the set of nodes and \REV{$E$} is the set of edges. Each node $i \in V$ corresponds to a control point $\boldsymbol{p_i}$. Each control point $\boldsymbol{p_i}$ is allowed to move in the $3D$ space, therefore, can be assigned a \REV{discrete} label $l_i \in \mathcal{L}$ associated to a displacement vector $\boldsymbol{d_{l_i}} \in \Re^3$. Notice that each $3D$ displacement vector is a tuple defined as $\boldsymbol{d_{l_i}} = (d_x, d_y, d_z)$, where $d_x$, $d_y$, and $d_z$ are the displacements in the $x$, $y$, and $z$ directions, respectively. A quantized version of the deformation field $\mathcal{D}^\Gamma$ is associated to a labeling $\Gamma$ of the graph $\mathcal{G}$. In another words, the labeling $\Gamma \in \mathcal{L}^{|V|}$ assign\REV{ed} to a each node $i \in V$ a label $l_i$ associated to a displacement vector $\boldsymbol{d_{l_i}}$. Note that $\mathcal{D}^\Gamma$ is a control point based representation of a dense deformation field, which is interpreted as a free form deformation model (FFD). \REV{FFDs were popularized in the medical image registration community by \cite{Rueckert1999}. They adopt a regular grid as parametric model and every control point contributes locally to the interpolation function when interpolating the dense deformation field.} We will call $\mathring{\mathcal{D}}^\Gamma$ to the dense deformation field obtained as a FFD interpolation of the control point based representation $\mathcal{D}^\Gamma$. We denote the control point $\boldsymbol{p'_i}$ as the new control point when the displacement $\boldsymbol{d_{l_i}}$ is applied to the original control point $\boldsymbol{p_i}$. Let us define a patch $\bar{\Omega}_{l_i}^I$ on the source image $I$ centered at the displaced control point $\boldsymbol{p'_i}$ (after applying label $l_i$). Similarly, we define $\Omega_i^J$ as the patch on the target image $J$ centered at the original control point $\boldsymbol{p_i}$, and $\bar{\Omega}_{l_i}^{S^I}$ as the patch on the input segmentation mask centered at the displaced control point $\boldsymbol{p'_i}$. 

\REV{At this point, it is important to make a distinction between the two different type of labels used in our framework. On one side, the semantic labels (or semantic classes) $c \in \mathcal{C}$ are provided as an input to our algorithm, defined at a pixel level, associated to anatomical structures and will be used to identify the context and choose the right combination of metrics. However, these labels are considered as input to the algorithm and therefore we do not predict them. On the other side, the labels $l_i \in \mathcal{L}$ are the latent variables (not provided as input), associated to displacement vectors and used to solve the registration problem by labeling the nodes $i \in V$ of the graph $G$.}
 
Given the definition of these variables, let us now define the unary feature vector corresponding to the $i^{th}$ node for a given label $l_i$ as $\mathcal{U}_i(l_i; I, J) = (u_1(\bar{\Omega}_{l_i}^I, \Omega_i^J), \cdots,  u_n(\bar{\Omega}_{l_i}^I, \Omega_i^J)) \in \mathbb{R}^n$, where $n$ is the number of metrics and $u_j(\bar{\Omega}_{l_i}^I, \Omega_i^J)$ is the unary feature corresponding to the $j^{th}$ metric evaluated using the patches $\bar{\Omega}_{l_i}^I$ and $\Omega_i^J$. For example, in case the $j^{th}$ metric is the mutual information (MI) then the unary feature $u_{MI}(\bar{\Omega}_{l_i}^I, \Omega_i^J)$ is the mutual information between the patches $\bar{\Omega}_{l_i}^I$ and $\Omega_i^J$.  In case of single metric, then $n$ is equal to 1. Recall that we have $|\mathcal{C}|$ number of \REV{semantic classes (anatomical structures)}. Therefore, given a weight matrix $W \in \mathbb{R}^{n \times |\mathcal{C}|}$, where $W(i,j)$ denotes the weight for the $i^{th}$ metric corresponding to the class $j$, the unary potential for the $i^{th}$ node for a given label $l_i$ is computed as follows:
\begin{align}
\label{eq:unaryPotential}
\bar{\mathcal{U}}_i(l_i; I, J, S^I, W) = {\bf w}(\bar{c})^\top \mathcal{U}_i(l_i; I, J) \in \mathbb{R},
\end{align}
where, ${\bf w}(\bar{c}) \in \mathbb{R}^n$ is the $\bar{c}^{th}$ column of the weight matrix $W$ and $\bar{c}$ is the most dominant class in the patch on the source segmentation mask $\bar{\Omega}_{l_i}^{S^I}$. The dominant class $\bar{c}$ is obtained as follows:
\begin{align}
\label{eq:dominantClass}
\bar{c} = \argmax_{c \in \mathcal{C}} f(\bar{\Omega}_{l_i}^{S^I}, c),
\end{align}
where, $f(\bar{\Omega}_{l_i}^{S^I}, c)$ is the number of voxels of class $c$ in the patch $\bar{\Omega}_{l_i}^{S^I}$ (see Figure \ref{fig:dominantClass} for a graphical example). Notice that one can use other criteria to find the dominant class. \REV{Also note that the semantic segmentation mask $S^I$ for the source image is given as an input to the registration framework, and is not imputed as a latent variable}.

\begin{algorithm}[t]
\caption{\REV{Deformable registration algorithm using a pyramidal approach in a discrete setting. Multiple grid levels are considered, from coarser to finer. The discrete displacement labels are initialized according to the grid size (method \textit{initializeLabelSpace}), restricting their size to
0.4 times the current grid spacing to guarantee the existence of the inverse for the final deformation field \cite{Glocker2011}. In every grid level, several iterations are performed by computing the optimal labeling and updating the current deformation field. In every iteration we further refine the displacement vectors by a factor of 0.7 (method \textit{refineLabelSpace}) to improve the sampling precision of the search space.}}
\begin{algorithmic}[1]
\Procedure{\REV{Register}}{\REV{$I$: Source, $J$: Target}}
\State \REV{$\mathring{\mathcal{D}}^\Gamma$  $\gets$ Initialize with a null deformation field}
\For{\REV{i=1 to gridLevels}}
\State \REV{$\mathcal{L} \gets initializeLabelSpace(i)$}
\For{\REV{j = 1 to iterationSteps}}
	  \State \REV{$\hat{\Gamma}$ $\gets$ Obtain optimal labeling solving Eq. (\ref{eq:registration})}
      \State \REV{$\mathring{\mathcal{D}}^\Gamma$ $\gets$ $updateDeformationField(\mathring{\mathcal{D}}^\Gamma, \hat{\Gamma}$)}
      \State \REV{$refineLabelSpace(\mathcal{L})$}
	\EndFor
	\EndFor
\State \REV{\textbf{return} $\mathring{\mathcal{D}}^\Gamma$ the final deformation field}
\EndProcedure
\end{algorithmic}
\label{algo:pyramidal}
\end{algorithm}

The pairwise clique potential between the control points $\boldsymbol{p_i}$ and $\boldsymbol{p_j}$ is defined as $f(l_i, l_j)$, where $f(\cdot, \cdot)$ is the $L_1$ norm between the two displacement vectors $\boldsymbol{d_{l_i}}, \boldsymbol{d_{l_j}}$ corresponding to the input labels (other regularizers could be used as well). \REV{Intuitively, this will encourage neighboring control points to be labeled with similar displacement vectors, enforcing smoothness in the final result.} Under this setting, the multi-class energy function corresponding to the deformable registration task is defined as:
\begin{equation}
\label{eq:energyFunctionFull}
	\mathcal{E}(\Gamma; I, J, S^I, W) = \sum_{i \in V} \bar{\mathcal{U}}_i(l_i; I, J, S^I, W) + \sum_{(i,j) \in E} f(l_i, l_j).
\end{equation}

Therefore, we aim at finding the optimal labeling $\hat{\Gamma}$ (associated to the quantized version of the deformation field ${\mathcal{D}}^{\hat{\Gamma}}$) by solving the following problem:
\begin{equation}
\label{eq:registration}
	\hat{\Gamma} = \argmin_{\Gamma \in \mathcal{L}^{|V|}} \mathcal{E}(\Gamma; I, J, S^I, W).
\end{equation}

\noindent \textbf{\REV{Pyramidal Approach}} Similar to~\cite{Glocker2008,ferrante2017slice}, we adopt a pyramidal approach \REV{(detailed in Algorithm \ref{algo:pyramidal})} to refine the search space at every level and, at the same time, capture a big range of deformations. \REV{The pyramidal approach consists in registering the images incrementally, at different resolutions (constructing a Gaussian pyramid of downsampled images) and associating to every level a different grid spacing. Coarser grids will be used to register the low resolution images, while finer grids will be used for the high resolution. The final deformation field is simply constructed through composition of the intermediate deformation fields. By combining the pyramidal approach with the L1 regularization we ensure smoothness on the deformation field while keeping a good capture range. In the multi-resolution strategy we restrict the size of the displacement vectors in the label space $\mathcal{L}$ to be 0.4 times the spacing between the control points of the grid. As suggested in \cite{Glocker2011}, this constraint in the size of the displacement vectors ensures that the final deformation field is diffeomorphic by construction, meaning that they are differentiable and invertible, and thus preserve topology.}

\noindent \textbf{\REV{Discrete Optimization}}
We use the well known FastPD~\cite{Komodakis2007} as the inference algorithm at every level of the pyramid. \REV{FastPD is a discrete optimization algorithm based on principles from linear programming and primal dual strategies, which applies the popular primal-dual schema to the relaxed version of the MRF integer programming formulation}.  Notice that the energy function from equation~\ref{eq:energyFunctionFull} is defined over the nodes and the edges of the sparse graph $G$ which represents the deformation field, and not over the dense voxels and the neighbourhood system defined over the input image $I$. The reason being that the input images are too big and thus can not be optimized efficiently. Once we obtain the optimal labeling $\hat{\Gamma}$, we estimate the dense deformation field $\mathring{\mathcal{D}}^{\hat{\Gamma}}$ from its quantized representation $\mathcal{D}^{\hat{\Gamma}}$ using the FFD interpolation model~\cite{Rueckert1999} in order to warp the input image, as it was already mentioned.

\section{Weakly-Supervised Learning of Metric Aggregations thorugh LSSVM}
\label{sec:learningParameters}
In the previous section we assumed that the weight matrix $W$ is given to us. However, this assumption becomes unrealistic quickly as the number of metrics and the classes increases. In order to circumvent this problem, we propose an algorithm to learn the weights using a given dataset. Our algorithm is based on the well known latent structured {\sc svm} framework~\cite{Taskar2003,Tsochantaridis2004,Yu2009} which optimizes an upperbound on the empirical risk. Instead of learning the complete weight matrix at once, we learn the weights for each class $c \in \mathcal{C}$ individually. From now onwards, the weight vector ${\bf w}_c$ denotes a particular column of the weight matrix $W$, which represents the weights corresponding to a particular class. We use the words `parameters' and `weights' interchangeably. In what follows we talk about the learning algorithm in details.

\subsection{Preliminaries}
\begin{figure}[t!]
  \centering
  \includegraphics[scale=0.4]{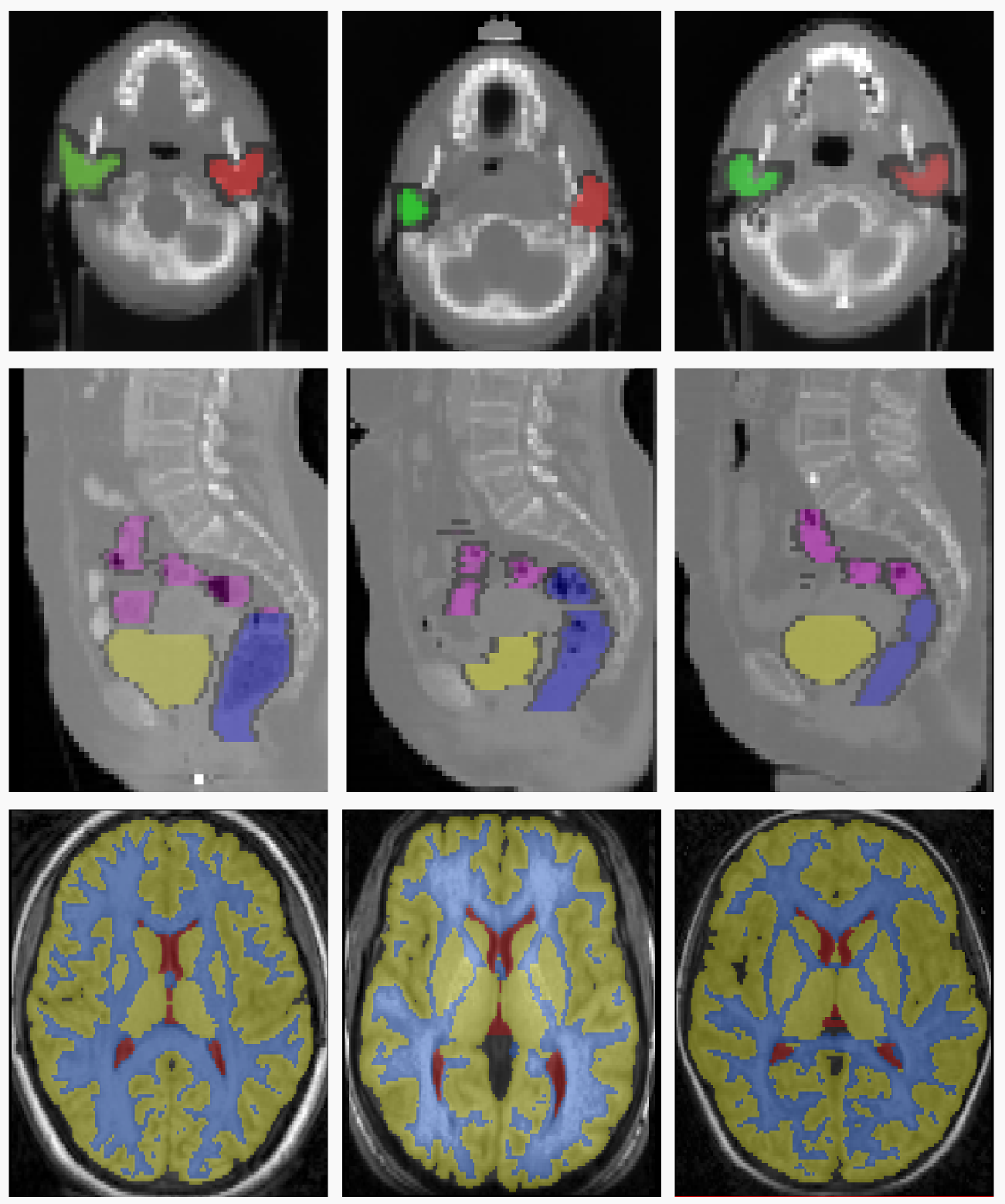}
  \caption{Sample slices from three different volumes of the RT Parotids, RT Abdominal and IBSR datasets.The top row represents the sample slices from three different volumes of the RT Parotids dataset. The middle row represents the sample slices of the RT Abdominal dataset, and the last row represents the sample slices from the IBSR dataset.} 
  \label{fig:imageExamples}
\end{figure}
We consider a dataset $D = \{( {\bf x}_i, {\bf y}_i )\}_{i=1,\cdots,N}$. Each ${\bf x}_i$ is a pair represented as ${\bf x}_i = (I_i, J_i)$, where $I_i$ is the source volume (or the source image) and $J_i$ is the target volume (or the target image). Similarly, each ${\bf y}_i$ is a pair represented as ${\bf y}_i = (S^I_i, S^J_i)$, where $S^I_i$ and $S^J_i$ are the segmentation masks for the source and target images, respectively. The size of each segmentation mask is the same as that of the corresponding images. As stated earlier, the segmentation mask is formed by the elements (or voxels) $s_k \in \mathcal{C}$, where $ \mathcal{C}$ is the set of classes. 

The loss function $\Delta(S^I, S^J) \in \mathbb{R}_{\geq 0}$ evaluates the similarity between the two segmentation masks $S^I$ and $S^J$. Higher value of $\Delta(., .)$ implies stronger dissimilarity between the segmentations. Since our final evaluation is based on the Dice coefficient, therefore, we use a Dice based loss function. Note that we follow the empirical risk minimization strategy, in which we should optimize the risk over which the evaluation is performed. The loss is therefore defined as:
\begin{equation} 
\label{eq:Loss}
\Delta(S^I, S^J)  = 1 - Dice(S^I, S^J).
\end{equation}
We approximate the Dice between the segmentation masks as defined below:
\begin{align}
Dice(S^I, S^J) = 2 \sum_{i \in V} \frac{|\phi(\boldsymbol{p_i^I}) \cap \phi(\boldsymbol{p_i^J})|}{|\phi(\boldsymbol{p_i^I})| + |\phi(\boldsymbol{p_i^J})|},
\end{align}
where, $\phi(\boldsymbol{p_i^I})$ and $\phi(\boldsymbol{p_i^J})$ are the patches at the control point $\boldsymbol{p_i}$ on the segmentation masks $S^I$ and $S^J$, respectively. The function $|.|$ represents the cardinality of a given set. The above approximation of the Dice makes it decomposable over the nodes of the graph $G$. As will be discussed shortly, this decomposition allows us to train our algorithm very efficiently.

Given the parameters ${\bf w}_c$ for a particular class, the labeling $\Gamma$, and the input tuple ${\bf x}$, the mutli-class energy function~(\ref{eq:energyFunctionFull}) can be trivially converted into class based energy function as follows:
\begin{equation}
\label{eq:energyFunctionPerClass}
	\mathcal{E}_c(\Gamma;{\bf x}, {\bf w}) = {\bf w}_c^\top \sum_{i \in V} \mathcal{U}_i(l_i; {\bf x}) + w_p \sum_{(i,j) \in E} f(l_i, l_j).
\end{equation}
where, $w_p \in \mathbb{R}_{\geq 0}$ is the parameter associated with the pairwise term. Let us denote the parameter vector ${\bf w} \in \mathbb{R}^{n+1}$ as the concatenation of ${\bf w}_c$ and $w_p$. The energy function~(\ref{eq:energyFunctionPerClass}) is linear in ${\bf w}$ and can be written as:
\begin{equation}
\label{eq:energyFunctionPsi}
	\mathcal{E}_c(\Gamma; {\bf x}, {\bf w}) = {\bf w}^\top \Psi(\Gamma; {\bf x}).
\end{equation}
where, $\Psi(\Gamma; {\bf x}) \in \mathbb{R}^{n+1}$ is the joint feature map defined as:
\begin{equation}
\label{eq:Psi}
	\Psi (\Gamma; {\bf x}) = 
	\begin{pmatrix} \sum_{i \in V} \mathcal{U}^1_i(l_i; {\bf x}) \\ \sum_{i \in V} \mathcal{U}^2_i(l_i; {\bf x}) \\ \vdots \\ \sum_{i \in V} \mathcal{U}^{n}_i(l_i; {\bf x}) \\ \sum_{(i,j) \in E} f(l_i, l_j) \\ \end{pmatrix}
\end{equation}\\
Notice that the energy function~(\ref{eq:energyFunctionPerClass}) does not depend on the source segmentation mask $S^I$. Source segmentation masks in the energy function~(\ref{eq:energyFunctionFull}) are used to obtain the dominant class using the equation~(\ref{eq:dominantClass}), which in this case is not required. However, we will shortly see that the source segmentation mask $S^I$ plays a crucial role in the learning algorithm to compute the loss function.

Ideally, the dataset $D$ must contain the ground truth deformation labeling $\Gamma$ corresponding to the source image $I$ in order to compute the energy term defined in the equation~(\ref{eq:energyFunctionPerClass}). Since annotating the dataset with the ground truth deformation is non-trivial, we use them as the latent variables in our algorithm. As will be seen shortly, we impute these deformations using the given dataset ensuring that the loss (as defined in the equation~(\ref{eq:Loss})) between the source segmentation mask when deformed using the imputed deformation field, and the target segmentation mask is minimized. In what follows, we will refer to this step indistinctly as 'latent assignment step' or 'segmentation consistent registration'.

\subsection{The Objective Function}
Given the dataset $D$, we would like to learn the parameter vector ${\bf w}$ such that minimizing the energy function~(\ref{eq:energyFunctionPerClass}) leads to a deformation field which when applied to the source segmentation mask gives minimum loss with respect to the target segmentation mask. Let us denote $S \circ \mathring{\mathcal{D}}^\Gamma$ the deformed segmentation when the dense deformation $\mathring{\mathcal{D}}^\Gamma$ is applied to the segmentation mask $S$. Therefore, ideally, we would like to learn ${\bf w}$ such that:
\begin{align}
\label{eq:empiricalRisk}
{\bf w}^* = \argmin_{\bf w} \frac{1}{N} \sum_i \Delta(S_i^I \circ \mathring{\mathcal{D}}^{\bar{\Gamma}}, S_i^J).
\end{align}
where, $\bar{\Gamma} = \argmin_{\Gamma} \mathcal{E}(\Gamma; {\bf x}_i, {\bf w})$. The above objective function is the empirical risk minimization based formulation. However, the objective function is highly non-convex in ${\bf w}$, therefore, minimizing it directly makes the algorithm sensitive in terms of convergence  to bad local minima. In order to circumvent this problem, we optimize a regularized upper bound on the loss as follows:
\begin{eqnarray}
\label{eq:learningObjective}
\min_{\bf w, \{\xi_i\}} && \frac{1}{2}||{\bf w}||^2 + \alpha ||{\bf w} - {\bf w}_0 ||^2 + \frac{C}{N} \sum_i \xi_i , \\
s.t. && \label{eq:constraintLSSVM} \min_{\Gamma,  \Delta(S_i^I \circ \mathring{\mathcal{D}}^\Gamma, S_i^J) =0} {\bf w}^\top \Psi({\bf x}_i, \Gamma) \leq {\bf w}^\top \Psi({\bf x}_i, \bar{\Gamma}) \nonumber - \\
&& \hspace{68px} \Delta(S_i^I \circ \mathring{\mathcal{D}}^{\bar{\Gamma}}, S_i^J) + \xi_i, \forall \bar{\Gamma}, \forall i \\
&& w_p \geq 0, \xi_i \geq 0, \forall i.
\end{eqnarray}
The above objective function minimizes an upper bound on the Dice based loss function denoted as the variable $\xi_i$, known as the slack. The first term in the objective function $||{\bf w}||^2$, is the regularization term used to avoid overfitting. The effect of the regularization term is controlled by the hyper-parameter $C$. The second term is the proximity term. This ensures that the learned ${\bf w}$ is close to the initialization ${\bf w}_0$. The effect of the proximity term can be controlled by the hyperparameter $\alpha$. \REV{The proximity term can be useful when we have some intuition about a good initialization (for example, if we know that certain 
factor works well when using the single metric approach, we could re-scale it taking into account the number of metrics in the multi-metric approach, and use it for initialization). In case we do not have any intuition, we can simply set $\alpha=0$ and this term will be ignored.}

Intuitively, for a given input sample, the constraints of the above objective function tries to enforce the condition that the energy corresponding to the best possible deformation field (with minimum loss, additional constraint to enforce coherence) must always be less than the energy corresponding to any other deformation field with a margin proportional to the loss. Notice that, the term $\min_{\Gamma,  \Delta(S_i^I \circ \mathring{\mathcal{D}}^\Gamma, S_i^J) =0} {\bf w}^\top \Psi({\bf x}_i, \Gamma)$ in equation~\ref{eq:constraintLSSVM}, makes the problem non convex. Shortly we will see how to upperbound this term, also known as the latent variable imputation step, in order to make the problem convex.

\subsection{The Learning Algorithm}
The objective function~(\ref{eq:learningObjective}) that optimizes an upper bound on the empirical risk, is non-convex. Hence, it can not be optimized efficiently to obtain the optimal set of parameters. However, it can be shown that the objective function is a non-convex function that can be re-written as the difference of convex functions, which can be seen as the sum of the convex and the concave functions. For such family of non-convex functions, the well known {\sc cccp} algorithm~\cite{Yuille2003}) can be used to obtain efficiently a local minima or a saddle point (the cutting plane algorithm can produce the optimal solution, but in practice this is not possible due to the computational constraints associated with the dimension of the problem). Broadly speaking, the {\sc cccp} algorithm consist of three steps --- (1) upperbounding the concave part at a given ${\bf w}$, which leads to an affine function in ${\bf w}$; (2) optimizing the resultant convex function (sum of convex and affine functions is convex); (3) repeating the above steps until the objective can not be decreased beyond a given tolerance of $\epsilon$. 

\begin{algorithm}[t!]
\caption{The {\sc cccp} algorithm (detailed version).}
\label{algo:cccpDetailed}
\fontsize{9}{9}\selectfont
\begin{algorithmic}[1]
\State $\mathcal{D}$, ${\bf w}_0$, $C$, $\alpha$, $\eta$, the tolerance $\epsilon$.
\State $t \leftarrow 0$.
\State ${\bf w}_t \leftarrow {\bf w}_0$
\Repeat
\State For a given ${\bf w}_t$, impute the latent variables $\hat{\Gamma}_i$ for each sample by solving the semantic consistent registration: 
		\begin{eqnarray}
		\hat{\Gamma}_i = \argmin_{\Gamma \in \mathcal{L}^{|V|}} \Big( {\bf w}_t^\top \Psi({\bf x}_i, \Gamma) + \eta \Delta(S_i^I \circ \mathring{\mathcal{D}}^\Gamma, S_i^J) \Big).  \nonumber
		\end{eqnarray}
		\State Initialize the constraint set for each sample: $\mathcal{W}_i \leftarrow \emptyset, \forall i$.
		\Repeat
		\State Obtain the most violated constraint (compute $\bar{\Gamma}_i$ for each sample):
		\begin{eqnarray}
		\bar{\Gamma}_i =  \argmin_{\bar{\Gamma} \in \mathcal{L}^{|V|}} \Big( {\bf w}^\top \Psi({\bf x}_i, \bar{\Gamma}) - \Delta(S_i^I \circ \mathring{\mathcal{D}}^{\bar{\Gamma_i}}, S_i^J) \Big).   \nonumber
		\end{eqnarray}		 
		 \State Update constraint set if $\bar{\Gamma}_i$ is sufficiently violated.
		 \begin{equation}
		  \mathcal{W}_i \leftarrow \mathcal{W}_i \cup \bar{\Gamma}_i, \forall i. \nonumber
		 \end{equation}
		 
		\State Solve the following optimization problem to obtain ${\bf w}$:
		
		\begin{eqnarray}
			\label{eq:Obj}
			\min_{\bf w, \{\xi_i\}} && \frac{1}{2}||{\bf w}||^2 + \alpha ||{\bf w} - {\bf w}_0 ||^2 + \frac{C}{N} \sum_i \xi_i , \nonumber \\
s.t. && {\bf w}^\top \Psi({\bf x}_i, \hat{\Gamma}_i) \leq {\bf w}^\top \Psi({\bf x}_i, \bar{\Gamma_i}) - \Delta(S_i^I \circ \mathring{\mathcal{D}}^{\bar{\Gamma}}, S_i^J) + \nonumber \\ 
&& \hspace{60px} \xi_i, \forall \bar{\Gamma_i} \in \mathcal{W}_i, \forall i \nonumber \\
&& w_p \geq 0, \xi_i \geq 0, \forall i.  \nonumber
		\end{eqnarray}
		\Until{No working set $\mathcal{W}_i$ can be further updated.}
		\State $t \leftarrow t+1$
		\State Update the parameters: ${\bf w}_t \leftarrow {\bf w}$
		\Until{Objective of the problem does not decrease more than $\epsilon$}.

    \end{algorithmic}
\end{algorithm}

The {\sc cccp} algorithm for the optimization of the objective function~(\ref{eq:learningObjective}) is shown in the Algorithm~\ref{algo:cccpDetailed}. The first step of upperbounding the concave functions (Algorithm~\ref{algo:cccpDetailed}, Line 5) is the same as the latent imputation step, which we call the {\em segmentation consistent registration} problem. The second step is the optimization of the resultant convex problem, which, in this case, is the optimization of the {\sc ssvm} (Algorithm~\ref{algo:cccpDetailed}, Lines 7 to 11)). The optimization leads to updating the parameters. We use the well known cutting plane algorithm~\cite{Joachims2009} for this purpose. In what follows we discuss these steps in detail.

\paragraph{Segmentation Consistent Registration} As discussed, the ground truth deformation is not known a priori. Thus, in this step, we generate the best possible ground truth deformation field at a given ${\bf w}$. This is same as the latent imputation step of the {\sc cccp} algorithm (Algorithm~\ref{algo:cccpDetailed}, Line 5). Recall that we are interested in learning the parameters ${\bf w}$ such that the upper bound on the loss function, defined in equation~(\ref{eq:Loss}), is minimized. This leads us to formulate the latent imputation step as an inference problem with additional constraints. These additional constraints ensure that the imputed deformation field deforms the input image such that the loss between the deformed input image and the target image is minimized. Mathematically, for a given parameter vector ${\bf w}$, the latent deformation is imputed by solving the following problem:
\begin{align}
\label{eq:imputeLatent}
\hat{\Gamma}_i = \argmin_{\Gamma \in \mathcal{L}^{|V|},  \Delta(S_i^I \circ \mathring{\mathcal{D}}^\Gamma, S_i^J) =0} {\bf w}^\top \Psi({\bf x}_i, \Gamma).
\end{align}
The above problem is hard to solve and may not have unique solution. Thus, we solve the relaxed version of the above problem as defined below:\begin{align}
\label{eq:imputeLatentRelaxed}
\hat{\Gamma}_i = \argmin_{\Gamma \in \mathcal{L}^{|V|}}  {\bf w}^\top \Psi({\bf x}_i, \Gamma) + \eta \Delta(S_i^I \circ \mathring{\mathcal{D}}^\Gamma, S_i^J). 
\end{align}
where, $\eta$ controls the relaxation trade-off parameter. Since the loss function used is decomposable, the above problem is equivalent to the inference of the deformable registration with simple modifications on the unary potentials. Thus, it can be solved efficiently using the FastPD inference algorithm.
\begin{figure}[t!]
  \centering
  \includegraphics[width=0.5\textwidth]{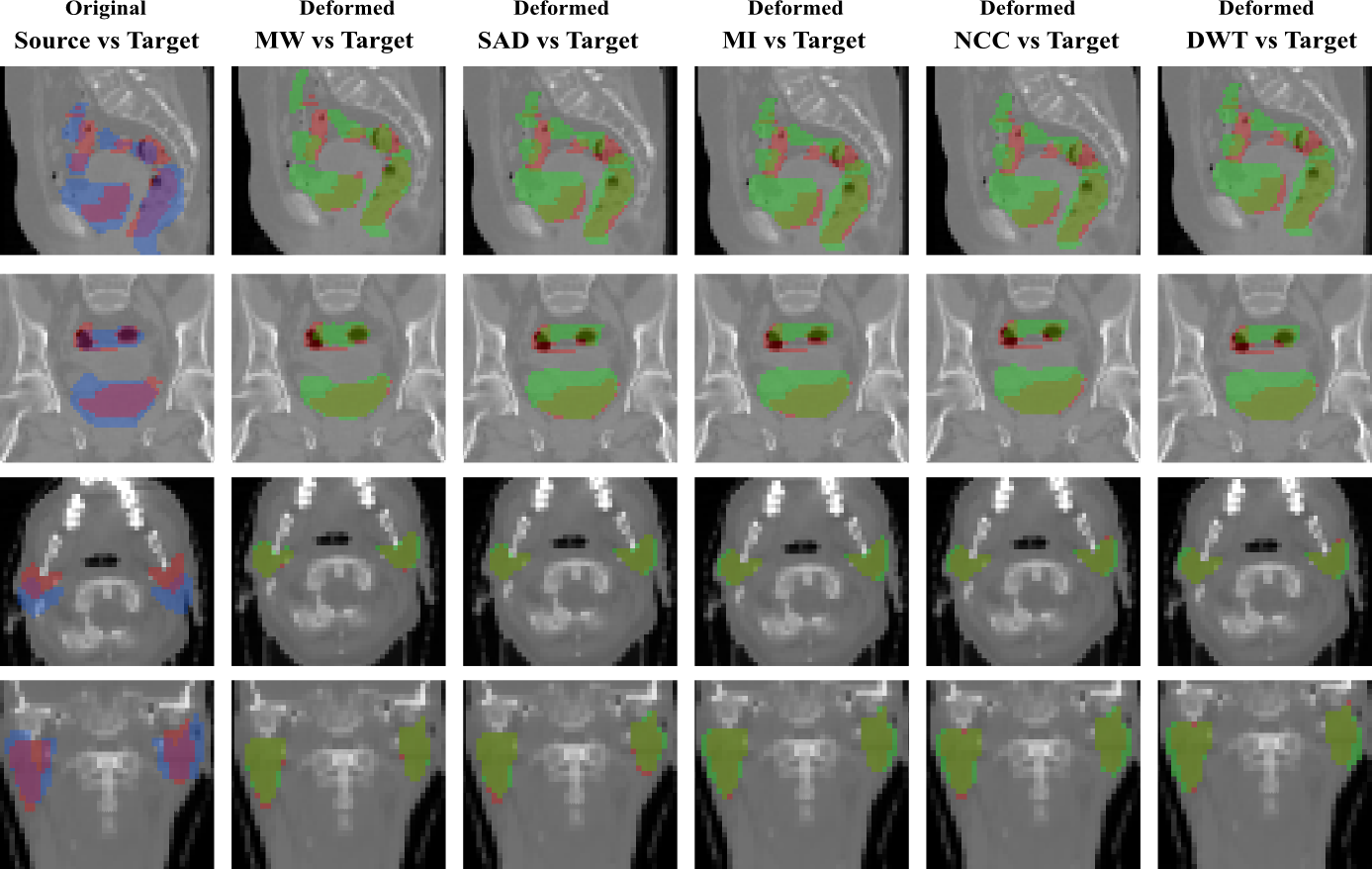}
  \caption{Overlapping of the segmentation masks in different views for one registration case from {\bf RT Abdominal} (first and second rows) and {\bf RT Parotids} (third and fourth rows) datasets. The first column corresponds to the overlapping before registration between the source (in blue) and target (in red) segmentation masks of the different anatomical structures of both datasets. From second to sixth column, we observe the overlapping between the warped source (in green) and the target (in red) segmentation masks, for the multi-weight algorithm (MW) and for the single metric algorithm using sum of absolute differences (SAD), mutual information (MI), normalized cross correlation (NCC) and discrete wavelet transform (DWT) as similarity measure.} 
  \label{fig:resultsAustriaLebanon}
\end{figure}

\paragraph{Updating the Parameters} Once the latent variables have been imputed or the concave functions have been upperbounded, the resultant objective function can be written as:
\begin{eqnarray}
\label{eq:learningObjectiveSSVM}
\min_{\bf w, \{\xi_i\}} && \frac{1}{2}||{\bf w}||^2 + \alpha ||{\bf w} - {\bf w}_0 ||^2 + \frac{C}{N} \sum_i \xi_i , \nonumber \\
s.t. && {\bf w}^\top \Psi({\bf x}_i, \hat{\Gamma}_i) \leq {\bf w}^\top \Psi({\bf x}_i, \bar{\Gamma}) - \\
&& \hspace{70px} \Delta(S_i^I \circ \mathring{\mathcal{D}}^{\bar{\Gamma}}, S_i^J) + \xi_i, \forall \bar{\Gamma}, \forall i \nonumber \\
&& w_p \geq 0, \xi_i \geq 0, \forall i.
\end{eqnarray}
where, $\hat{\Gamma}_i$ is the labeling associated to the quantized latent deformation field $\mathcal{D}^{\hat{\Gamma}_i}$ imputed by solving the problem~(\ref{eq:imputeLatentRelaxed}). Intuitively, the above objective function tries to learn the parameters ${\bf w}$ such that the energy corresponding to the imputed deformation field is always less than the energy of any other deformation field with a margin proportional to the loss function with some positive slack. Notice that the above objective function has exponential number of constraints, one for each possible labeling $\bar{\Gamma} \in \mathcal{L}^{|V|}$. In order to alleviate this problem we use the cutting plane algorithm~\cite{Joachims2009}. Let us briefly talk about the idea behind the cutting plane algorithm. For a given ${\bf w}$, each labeling $\bar{\Gamma}$ gives a slack. Therefore, instead of minimizing all the slacks for a particular sample at once, we rather find the labeling that leads to the maximum value of the slack and store this in a set known as the working set. This is known as finding the most violated constraint. Therefore, instead of using exponentially many constraints, we now optimize our algorithm over the constraints stored in the working set. This process is repeated till no constraints can be added to the working set. The main ingredient of the above discussed cutting plane algorithm is {\em finding the most violated constraint}. As discussed earlier, the most violated constraint for the $i^{th}$ sample is the deformation field associated to the labeling that maximizes the slack corresponding to this sample. Rearranging the terms in the constraints of the objective function~(\ref{eq:learningObjectiveSSVM}) to obtain the slack, ignoring the constant term ${\bf w}^\top \Psi({\bf x}_i, \hat{\Gamma}_i)$, and maximizing it with respect to the possible deformations (which is equivalent to minimizing the negative of it), leads to the following problem solving which gives the most violated constraint:
\begin{align}
\label{eq:mostViolated}
\bar{\Gamma}_i =  \argmin_{\Gamma \in \mathcal{L}^{|V|}} \Big( {\bf w}^\top \Psi({\bf x}_i, \bar{\Gamma}) - \Delta(S_i^I \circ \mathring{\mathcal{D}}^{\bar{\Gamma}}, S_i^J) \Big).
\end{align}
Since the loss function is decomposable, again, the above problem is equivalent to the deformable registration with modifications on the unary potentials. Thus, it can be solved efficiently using the FastPD.

\subsection{Prediction}
Once we obtain the learned parameters ${\bf w}_c$ for each class  $c \in \mathcal{C}$ using Algorithm~\ref{algo:cccpDetailed}, we form the matrix $W$ where each column of the matrix represents the learned parameter for a specific class. This matrix is then used to solve the registration problem defined in equation~(\ref{eq:energyFunctionFull}) using the approximate inference algorithm in a new setting where semantic labels are available in the source image but not in the target. \REV{Note that we use the pyramidal approach as described in Algorithm \ref{algo:pyramidal} at both training and prediction time.}

\begin{figure}[t!]
  \centering
  \includegraphics[width=0.5\textwidth]{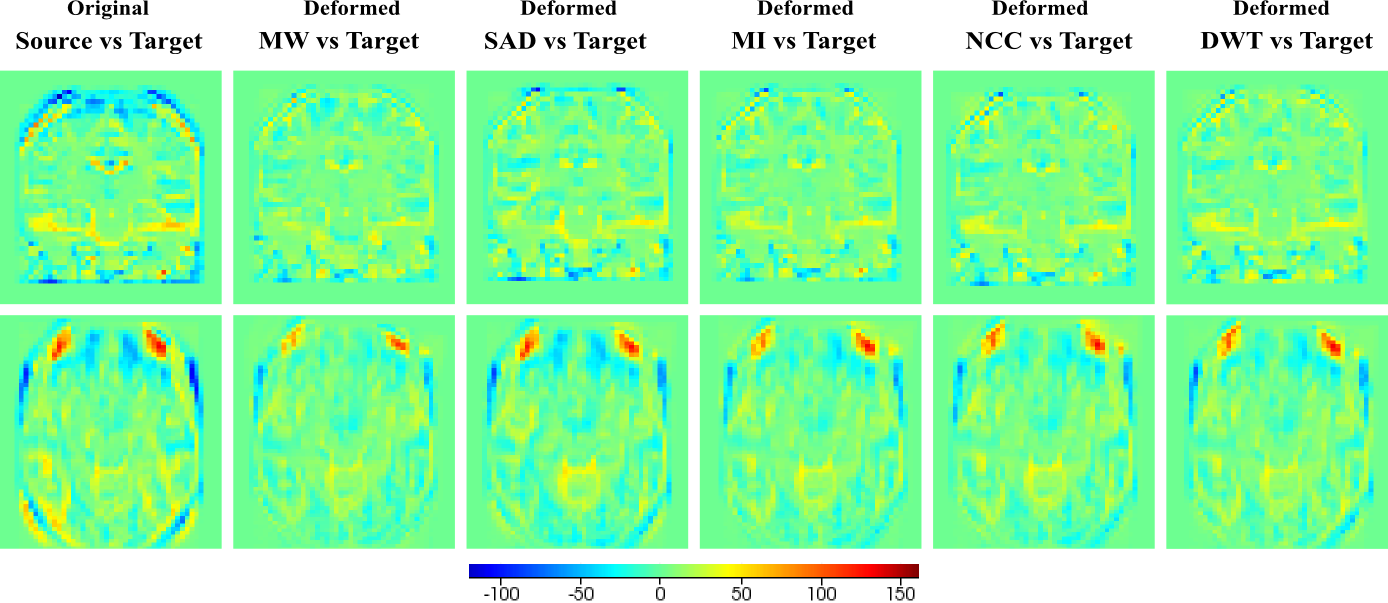}
  \caption[Qualitative results for one slice of one registration case from IBSR dataset.]{Qualitative results for one slice of one registration case from {\bf IBSR dataset}. In this case, since showing overlapped structures in the same image is too ambiguous given that the segmentation masks almost cover the complete image, we are showing the intensity difference between the two volumes. This is possible since images are coming from the same modality and they are normalized. The first column shows the difference of the original volumes before registration. From second to sixth column we observe the difference between the warped source and the target images, for the multi-weight algorithm (MW) and the single metric algorithm using sum of absolute differences (SAD), mutual information (MI), normalized cross crorrelation (NCC) and discrete wavelet transform (DWT) as similarity measure.} 
  \label{fig:resultsIBSRExtra}
\end{figure}

\section{Experiments and Results}
\label{sec:experiments}

We evaluated our method on three different medical datasets. These datasets involve several anatomical structures, different image modalities, and inter/intra patient images, which makes the deformable registration task on these dataset highly challenging. In all the experiments, we cross validate the hyper parameters $C$ and $\alpha$, and use $\eta=50$. We use four different metrics in all the experiments: (1) sum of absolute differences ({\sc sad}), (2) mutual information ({\sc mi}), (3) normalized cross correlation ({\sc ncc}), and (4) discrete wavelet coefficients ({\sc dwt}) (see \cite{singh2015quasi} for a complete description of DWT as a similarity measure for image registration). In all the experiments (single and multi-metric) we used the same set of parameters for the pyramidal approach based inference algorithm (discussed in the section~\ref{sec:regProblem}). These parameters are as follows: 2 pyramid levels, 5 refinement steps per pyramid level, 125 labels, and distance between control points of 25mm in the finer level. The running time for each registration case is around 12 seconds. For the training, we initialized ${\bf w}_0$ with the hand tuned values for each metric: ${\bf w}_0 = (0.1, 10, 10, 10)$, for {\sc sad}, {\sc mi}, {\sc ncc}, and {\sc dwt}, respectively. 

\REV{For comparison, we include results computed using two of the most popular and developed state-of-the-art registration toolboxes: Elastix \cite{klein2010elastix} and Advanced Normalization Tools (ANTs) \cite{avants2014insight}. Elastix implements a standard multi-resolution approach using stochastic gradient descent for optimization, b-spline as deformation model and the MI metric\footnote{\REV{The detailed parameters file used in our experiments can be found here: http://elastix.bigr.nl/wiki/images/a/ad/Par0000bspline.txt}}. For ANTs, we used the symmetric diffeomorphic  (SyN) implementation that optimizes the CC metric\footnote{\REV{We use the antsRegistrationSyN.sh script provided with the toolbox with the following parameters: "-d 3 -f fixedImg.nii.gz -m movingImg.nii.gz -o out".}}} Below we give details about the different datasets.

\begin{figure}[t!]
  \centering
  \includegraphics[scale=0.4]{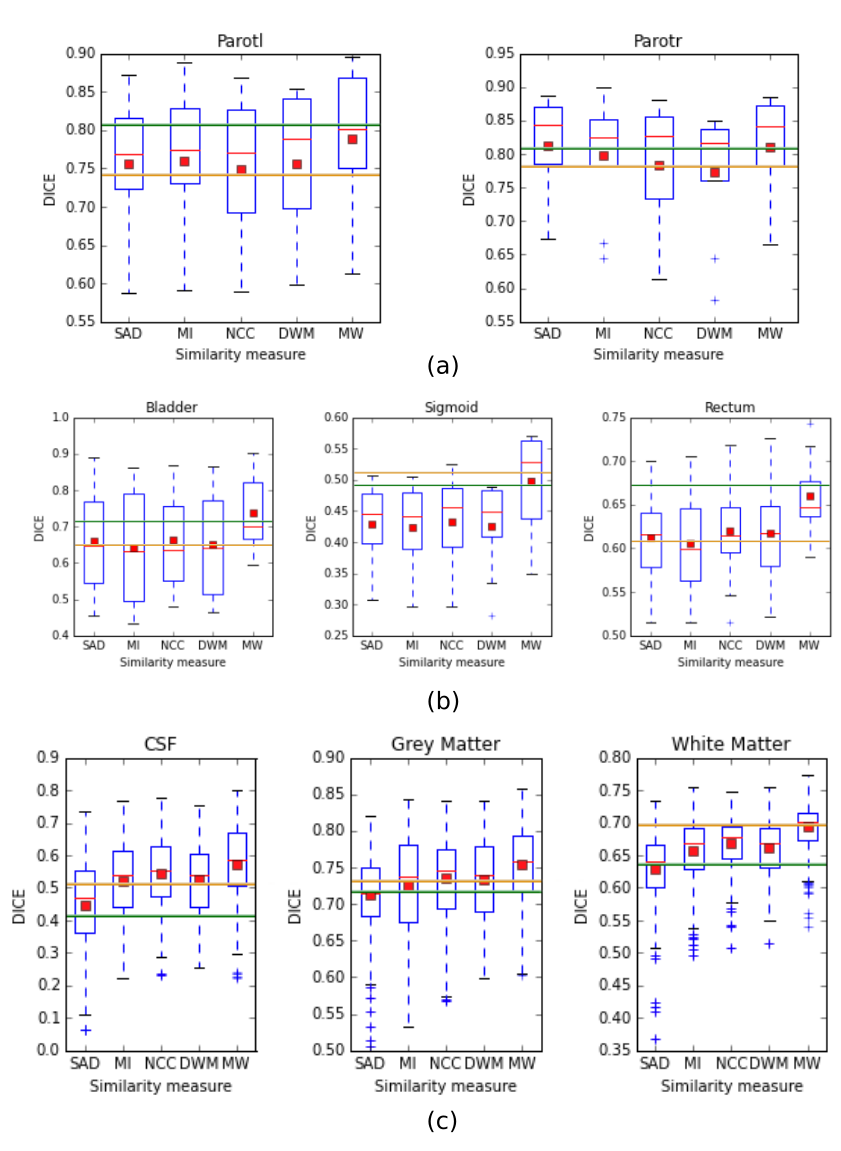}
  \caption{Results for the RT Parotids (a), RT Abdominal (b) and IBSR (c) datasets for the single-metric registration ({\sc sad}, {\sc mi}, {\sc ncc}, {\sc dwt}) and the multi-metric registration ({\sc mw}). The weights for the multi-metric registration are learned using the framework proposed in this work. The red square is the mean and the red bar is the median. It is clear from the results that using the learned linear combination of the metrics outperforms the single-metric based registration. \REV{Two baselines are included to confirm that the proposed method achieves state of the art accuracy: the orange line corresponds to Elastix and the green line to ANTs SyN.}} 
  \label{fig:resultsDice}
\end{figure}

\noindent \textbf{RT Parotids} It contains $8$ CT volumes of head, obtained from 4 different patients, 2 volumes per patient. The volumes are captured in two different stages of a radiotherapy treatment in order to estimate the radiation dose. Right and left parotid glands were segmented by the specialists in every volume. The dimensions of the volumes are $56\times62\times53$ voxels with a physical spacing of $3.45$mm, $3.45$mm, and $4$mm, in x, y, and z axes, respectively. We generated $8$ pairs of source and target volumes using the given dataset. Notice that, while generating the source and target pairs, we did not mix the volumes coming from different patients. We split the dataset into train and test, and cross validated the hyper-parameters $C$ and $\alpha$ on the train dataset. The average result on the test dataset are shown in the Figure~\ref{fig:resultsDice}.a, while qualitative results can be found in Figure \ref{fig:resultsAustriaLebanon}. 

\noindent \textbf{RT Abdominal} This dataset contains $5$ CT volumes of abdomen for a particular patient captured with a time window of about 7 days during a radiotherapy treatment. Three organs have been manually segmented by the specialists: (1) sigmoid, (2) rectum, and (3) bladder. The dimensions of the volumes are $90\times60\times80$ voxels with a physical spacing of $3.67$mm, $3.67$mm, and $4$mm, in x, y, and z axes, respectively (there are small variations depending on the volume). We generated a train dataset of 6 pairs and test dataset of 4 pairs. The results on the test dataset are shown in the Figure~\ref{fig:resultsDice}.b. 
\noindent \textbf{IBSR} We used images from the well known Internet Brain Segmentation Repository dataset, which consists of $18$ brain {\sc mri} volumes, coming from different patients. Segmentations of three different brain structures are provided: white mater (WM), gray mater (GM), and cerebrospinal fluid (CSF). We used a downsampled version of the dataset to reduce the computation. The dimension of the volumes are $64\times64\times64$ voxels with a physical spacing of $3.75$mm, $3.75$mm, and $3$mm in x, y, and z axes, respectively. To perform the experiments, we divided the 18 volumes in 2 folds of 9 volumes on each fold. This gave a total of 72 pairs per fold. We used an stochastic approach for the learning process, where we sample 10 different pairs from the training set, and we tested on the 72 pairs of the other fold. We run this experiment 3 times per fold, giving a total of 6 different experiments, with 72 testing samples and 10 training samples randomly chosen. Quantitative results on the test dataset are shown in Figure~\ref{fig:resultsDice}.c while qualitative results can be found in Figure \ref{fig:resultsIBSRExtra}. \\

As observed in Figure~\ref{fig:resultsDice}, the linear combination of similarity measures weighted using the learned coefficients outperforms the single metric based registration, \REV{improving the results of the discrete registration framework making them comparable (and in most of the cases better) than those obtained with the baselines (green and orange lines in Figure~\ref{fig:resultsDice})}. In all the cases the Dice for the multi-metric is higher than the Dice for the single metric based registrations, or it is as good as the best them (please refer to Figure~\ref{fig:resultsDice}.a, `Parotr' to see the case in which the multi-metric is at least as good as the best obtained using the single metric). The results for the `Sigmoid' organ in the Figure~\ref{fig:resultsDice}.b show that in some cases the multi-metric based registration can significantly outperform the single metric based registration. Table~\ref{tab:averageDiceImprovement} shows the average Dice value per organ for the three datasets, considering the single and multi-metric approaches. We achieved maximum average improvement of 0.033, 0.082 and 0.037 in terms of Dice coefficient for RT Parotids, RT Abdominal and IBSR. 

Figure~\ref{fig:imageExamples} shows the examples of the slices from the $3D$ volumes corresponding to each dataset. In figures \ref{fig:resultsAustriaLebanon} and \ref{fig:resultsIBSRExtra} we include some qualitative results on the three challenging datasets in order to highlight the effects of learning the weights of different metrics. In the first one (figure \ref{fig:resultsAustriaLebanon}), we present the overlapping of the segmentation masks in different views for one registration case from {\bf RT Abdominal} and {\bf RT Parotids} datasets, using single and multi-metric approaches. The observed results are coherent with the numerical results reported in figures~\ref{fig:resultsDice}.a and b. We observe that multi-weight algorithm gives a better fit between the deformed and ground truth structures than the rest of the single similarity measures, which are over segmenting most of the structures showing a poorer registration performance. In the second graph (figure \ref{fig:resultsIBSRExtra}), we include results for the IBSR dataset. Extreme values (which mean high differences between the images) correspond to blue and red colors, while green indicates no difference in terms of intensity. Note how most of the big differences observed in the first column (before registration) are reduced by the multi-weight algorithm, while some of them (specially in the peripheral area of the head) remain when using single metrics.

\begin{table}[t]
\centering
\caption{Average Dice value per organ, for the single and multi-metric approaches, are reported for the three datasets. The last column indicates the average Dice improvement that our proposed method provides when compared with the single metric approaches. We can observe improvements of a maximum of 8\% points in terms of Dice coefficient.}
\resizebox{\columnwidth}{!}{%
\begin{tabular}{|
>{\columncolor[HTML]{EFEFEF}}c |
>{\columncolor[HTML]{EFEFEF}}c |c|c|c|c|c|c|}
\hline
\cellcolor[HTML]{C0C0C0}\textbf{Dataset}               & \cellcolor[HTML]{C0C0C0}\textbf{Organ} & \cellcolor[HTML]{C0C0C0}\textbf{SAD} & \cellcolor[HTML]{C0C0C0}\textbf{MI} & \cellcolor[HTML]{C0C0C0}\textbf{NCC} & \cellcolor[HTML]{C0C0C0}\textbf{DWT} & \cellcolor[HTML]{C0C0C0}\textbf{MW} & \cellcolor[HTML]{C0C0C0}\textbf{Average Dice increment for MW} \\ \hline
\cellcolor[HTML]{EFEFEF}                               & Parotl                                 & 0,756                                & 0,760                               & 0,750                                & 0,757                                & 0,788                               & 0,033                                                          \\ \cline{2-8} 
\multirow{-2}{*}{\cellcolor[HTML]{EFEFEF}RT Parotids}  & Parotr                                 & 0,813                                & 0,798                               & 0,783                                & 0,774                                & 0,811                               & 0,019                                                          \\ \hline
\cellcolor[HTML]{EFEFEF}                               & Bladder                                & 0,661                                & 0,643                               & 0,662                                & 0,652                                & 0,736                               & 0,082                                                          \\ \cline{2-8} 
\cellcolor[HTML]{EFEFEF}                               & Sigmoid                                & 0,429                                & 0,423                               & 0,432                                & 0,426                                & 0,497                               & 0,070                                                          \\ \cline{2-8} 
\multirow{-3}{*}{\cellcolor[HTML]{EFEFEF}RT Abdominal} & Rectum                                 & 0,613                                & 0,606                               & 0,620                                & 0,617                                & 0,660                               & 0,046                                                          \\ \hline
\cellcolor[HTML]{EFEFEF}                               & CSF                                    & 0,447                                & 0,520                               & 0,543                                & 0,527                                & 0,546                               & 0,037                                                          \\ \cline{2-8} 
\cellcolor[HTML]{EFEFEF}                               & GM                                     & 0,712                                & 0,725                               & 0,735                                & 0,734                                & 0,761                               & 0,035                                                          \\ \cline{2-8} 
\multirow{-3}{*}{\cellcolor[HTML]{EFEFEF}IBSR}         & WM                                     & 0,629                                & 0,658                               & 0,669                                & 0,661                                & 0,682                               & 0,028                                                          \\ \hline
\end{tabular}
}

\label{tab:averageDiceImprovement}

\end{table}

\begin{figure*}[t!]
  \centering
  \includegraphics[width=1\textwidth]{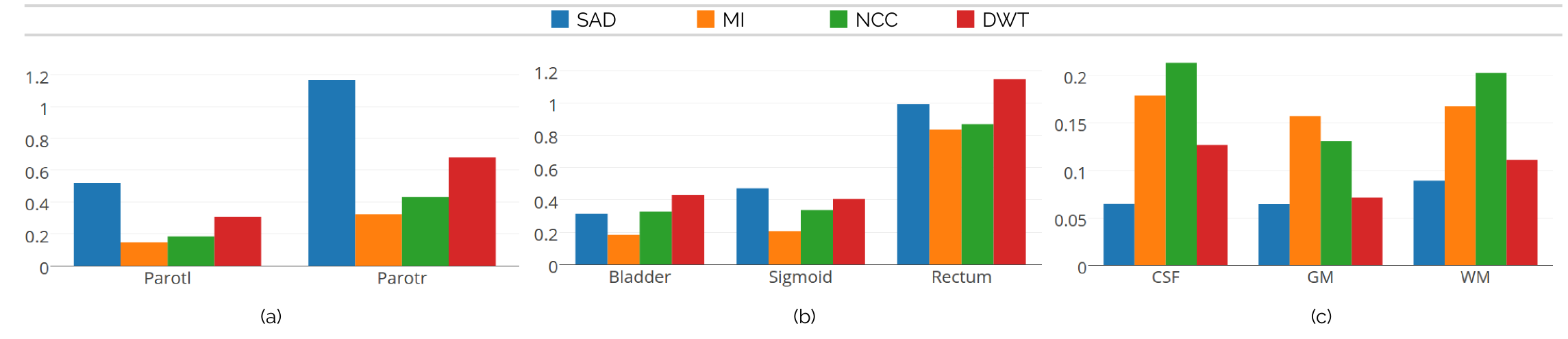}
  \caption{Example of learned weights for RT Parotids (a), RT Abdominal (b) and IBSR (c) datasets. Since the structures of interest in every dataset present different intensity distributions, different metric aggregations are learned.  } 
  \label{fig:learnedWeights}
\end{figure*}

\section{Discussion and Conclusions}
In this paper we introduced a novel framework to learn aggregations of similarity measures in the context of deformable image registration. We also proposed a multi-metric MRF based image registration algorithm that incorporates such metric aggregations by weighting different similarity measures depending on the anatomical regions. We showed that associating different similarity criteria to every anatomical region yields results superior to the classic single metric approach. In order to learn this mapping in real scenarios where ground truth is generally given in the form of segmentation masks, we proposed to conceive deformation fields as latent variables and solve our problem using the {\sc lssvm} framework. 

\REV{One of the main} limitations of our method is the need of segmentation masks for the source images at testing time. However, different real scenarios like radiation therapy or atlas-based segmentation methods fulfill this condition and can be improved through this technique. Note that, at prediction (testing) time, the segmentation mask is simply considered to determine the combination of metric weights per control node (as indicated in equation~\ref{eq:dominantClass} and Figure \ref{fig:dominantClass}). The segmentation labels are not used explicitly at testing time to guide the registration process that is purely image based. Instead, segmentation masks are required at test time just for the source image and only used to choose the metric aggregation learned for each dominant class. At training time though, both source and target segmentation mask are used to compute the loss function $\Delta$. The idea could be further extended to unlabeled data (as it concerns the source image at testing time) where the dominant label class per control node is the output of a classification/learning method.
\REV{Another limitation is that the metric aggregations are organ specific and therefore they must be learned whenever the framework is used in a new type of data. However, in real clinical scenarios the organs, image modalities and imaging machines considered in a given medical center will remain constant. Thus, the training has to be performed only once, when initially setting up the imaging pipeline, but this will not imply re-training for future patients.}

Figure \ref{fig:learnedWeights} shows the optimal weights learned for the three datasets. Note that in case of the RT Parotids, given that both parotid glands present the same intensity distribution, similar weights are learned for both structures, with SAD dominating the other similarity measures. However, in IBSR dataset, NCC dominates in case of CSF and WM, while MI receives the higher value for gray matter (GM). This is important since it indicates that different similarity measures complement each other, and suggests that the proposed learning framework takes advantage of this fact by assigning different weights depending on the anatomical structures.
%

From a theoretical viewpoint, we showed how three of the main components of LSSVM can be reduced to equivalent inference problems. In other words, the latent imputation step (Eq. \ref{eq:imputeLatentRelaxed}), the prediction step (Eq. \ref{eq:registration}) and finding the most violated constraint (Eq. \ref{eq:mostViolated}) can be formulated as the exact same problem. Since the loss function $\Delta$ was defined so that it distributes with respect to the unary terms of the energy, the only difference among these three problems will be given by the unary potentials. This is extremely important given that further improvements in this inference problem will directly increase the quality of the results.

Rather than employing highly customized solutions that suffer from robustness and modularity, our approach relies on producing class-dependent metrics as linear combination of widely known and conventional mono-modal and multi-modal metrics. Consequently, the proposed registration method is modular and might be adjusted to different setting by simply changing the linear weights. Extensive experimental validation on various challenging datasets demonstrated the potentials of the proposed method.

The registration weights were learned by minimizing an upperbound on the Dice based loss function. Dice is conventionally used but does not offer a very convincing picture as it concerns registration performance. The integration of alternative accuracy measures such as the Hausdorff distance between surfaces or even real geometric distances for anatomical landmarks could further enhance the performance and the robustness of the method. 
\REV{The use of alternative parameter learning mechanisms is another interesting approach to explore. In that sense, learning a combination of metrics could be seen as a multi-view learning problem, where every metric is considered as a view of the original image. We plan to explore the use of several novel frameworks to deal with multi-view learning \cite{luo2015tensor,luo2016large} in the context of learning metric aggregations.} Last but not least, the use of the method on clinical applications where domain knowledge is present (for example, radiotherapy) could be considered to improve existing patient positioning practices.  

\section*{Acknowledgment}
EF is beneficiary of an AXA Research Fund grant. 

\bibliographystyle{ieeetr}
\bibliography{library}

\end{document}